\documentclass[letterpaper, 10 pt, conference]{ieeeconf}  

\IEEEoverridecommandlockouts                              

\usepackage{cite}
\usepackage{amsmath,amssymb,amsfonts}
\usepackage{graphicx}
\usepackage{textcomp}
\usepackage{xcolor}
\usepackage{booktabs}
\usepackage{array}
\usepackage{multirow}
\usepackage{algorithm}
\usepackage{algpseudocode}

\title{\LARGE \bf
Memory Group Sampling Based Online Action Recognition Using Kinetic Skeleton Features*
}

\author{Guoliang Liu, Qinghui Zhang, Yichao Cao, Junwei Li, Hao Wu and Guohui Tian 
\thanks{*This work is supported by the National Key R$\&$D Program of China under grant 2018YFB1306500,
National Natural Science Foundation of China under grant 91748115 and 61603213,
Young Scholars Program of Shandong University under grant 2018WLJH71,
the Fundamental Research Funds of Shandong University,
and the Taishan Scholars Program of Shandong Province.}
\thanks{ Guoliang Liu, Qinghui Zhang, Yichao Cao, Junwei Li, Hao Wu and Guohui Tian are with School of Control Science and Engineering, Shandong University, 250061 Jinan, China.
        {\tt\small liuguoliang@sdu.edu.cn}}%
}

\begin{document}

\maketitle
\thispagestyle{empty}
\pagestyle{empty}

\begin{abstract}

Online action recognition is an important task for human centered intelligent services,  
which is still difficult to achieve due to the varieties and uncertainties of spatial and temporal scales of human actions.
In this paper, we propose two core ideas to handle the online action recognition problem. 
First, we combine the spatial and temporal skeleton features to depict the actions, which include not only the geometrical features, but also multi-scale motion features, such that both the spatial and temporal information of the action are covered. 
Second, we propose a memory group sampling method to combine the previous action frames and current action frames, which is based on the truth that the neighbouring frames are largely redundant, and the sampling mechanism ensures that the long-term contextual information is also considered. 
Finally, an improved 1D CNN network is employed for training and testing using the features from sampled frames. 
The comparison results to the state of the art methods using the public datasets show that the proposed method is fast and efficient, and has competitive performance.
\end{abstract}

\section{INTRODUCTION}

Online action recognition has important application value in elderly care, medical rehabilitation, security surveillance, and human-robot interaction and collaboration. 
There are many sensors can be used to capture human actions, e.g., RGB camera, RGBD camera, IMU, 3D laser scanner. 
The RGB camera is the most common sensor to analyze human actions \cite{donahue2015long} \cite{zhang2012spatio}, since it is similar to the human biology vision that can capture natural and substantial informations. In addition, its price is more acceptable than IMU and laser scanners. However, the 2D image is projected from 3D world, such that the human action recognition can be affected by the view point. In contrast, the recent popular RGBD camera can directly output the depth information of the environment, which can be used for 3D human skeleton pose detection \cite{zhu2016co} \cite{du2015hierarchical} that is invariant to the view direction, e.g., Microsoft Kinect, Intel realsense, Asus Xtion. In addition, the RGBD camera has low cost price, real time 3D reconstruction capability and easy-to-use features, which drive its popularity in the field of human pose estimation and action recognition.

The skeleton of the human is one of the most common way to model human pose using RGBD sensors, which includes a number of spatial connected bone joints. Each joint has a 2D coordinate in the RGB image or depth image, such that the skeleton is just a vector with very limited length, which can save computational cost to analyze the human pose and action. Currently, most of the human action recognition methods using skeleton data are offline processing, which use a recorded video clip with a fixed length as input \cite{kong2017deep}, or predict the start frame and end frame of the action in the sequential images \cite{gao2017turn} \cite{lea2017temporal}. On the other hand, the online action recognition is still an onging problem which is difficult to be solved, since only previous frames of current time are available and the start point of the action is unknown \cite{li2016online} \cite{baek2017real}. Furthermore, the real time requirement of the online action recognition is the other challenge issue due to the expensive computational cost of the recent deep learning algorithms.   

In this paper, we focus on online action recognition using 3D skeleton sequences derived from RGBD sensors. The novelty of our idea are as follows: 

(1) We first introduces a group of kinetic skeleton features that can capture both of the spatial and temporal features of the human action, which includes joint collection distance feature, multi-scale motion feature and geometrical features. 

(2) We propose to use a memory group sampling mechanism to handle the uncertainty of the temporal scale of the actions, such that long-term contextual information can be considered for action recognition. 

(3) We then use a fast, small and efficient neural network to combine these kinetic skeleton features, which can return a competitive performance for online action recognition. 

The rest parts of the paper are structured as follows. We first discuss related works in the field of action recognition using skeleton data in Section \ref{sec:relate}, and then introduce our kinetic skeleton features, memory group sampling and neural network for action recognition in Section \ref{sec:method}. The demonstrated experiments on the public datasets can be found in Section \ref{sec:exp}. Finally, the paper is concluded in Section \ref{sec:conc}.

\begin{figure*}
\centering
\includegraphics[width=\textwidth]{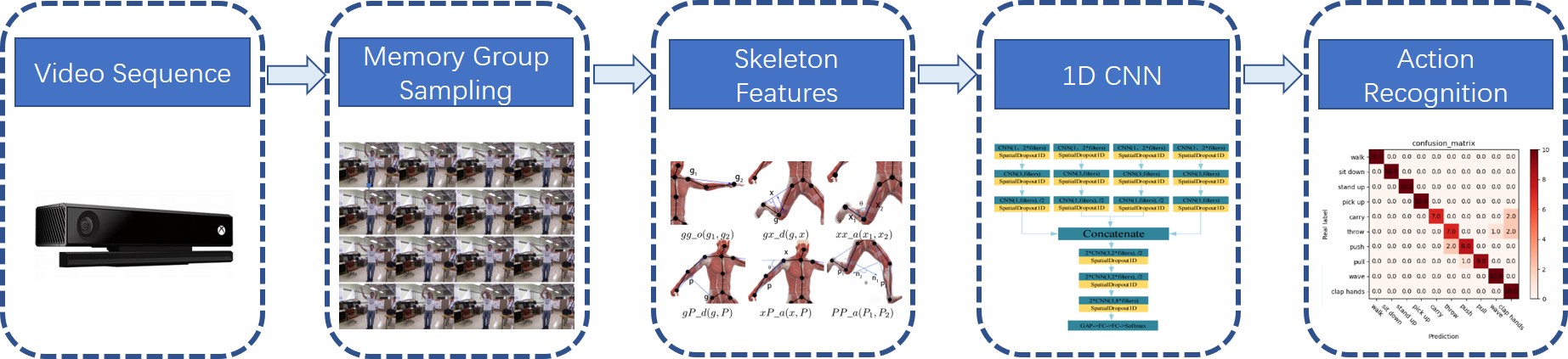}\\
\caption{The overall flowchart of the proposed online human action recognition method using memory group sampling with advanced geometry and motion skeleton features. }
\label{fig:flowchart}
\end{figure*}

\section{Related Work} \label{sec:relate}
In this section, we review the related works of action recognition using skeleton data, which includes offline algorithms and online algorithms.

\subsection{Offline Action Recognition}

For offline action recognition, the skeleton sequence data for processing is often segmented for each action in advance, such that we can easily label these data and train the learning methods. In addition, it is convenient to use segmented data for numerical accuracy analysis. Furthermore, the computation cost of the algorithm is not key problem for the offline processing. H. Wang and L. Wang \cite{wang2017modeling} proposed a new two-stream RNN architecture to model the temporal dynamics and spatial structure of skeleton for action recognition. In order to improve the generalization ability of the model, they further developed a data expansion technology based on 3D transformation, including rotation and scaling transformation. Finally, the recognition accuracy of the algorithm is verified on two public datasets. Liu et al. \cite{liu2016spatio} proposed a LSTM model based action recognition method, which simultaneously extracts the action-related information in the temporal and spacial domains. Wang et al. \cite{wang2016action} proposed the Joint Trajectory Maps (JTM), which represents spatial configuration and dynamics of joint trajectories into three texture images through color encoding. Then they use CNN for action recognition. The original skeleton data can be affected by the view direction, so more advanced geometrical features can be extracted from original joints. For instance, Zhang et al. \cite{zhang2017geometric} designed eight geometric features to represent raw skeleton data, including joint-to-joint distances, joint-line distances, joint-to-plane distances, etc. These hand-crafted features are used as the input of LSTM network for action recognition. They finally proved that properly defining hand-crafted features for a basic model can be superior. 

\subsection{Online Action Recognition}

Online action recognition refers to predict the category of ongoing action based on the observed data up to the present, which means we need to predict the category of action before they are completely executed. The works for online action recognition is fewer than the offline algorithms. Most of current online algorithms mainly deal with RGB videos. The method proposed by Geest et al. \cite{de2016online} took a RGB video stream as input, and outputs the class of action in real time. One of the challenge problems of online action recognition is the unknown starting point of the action, a sliding window with fixed scale proposed by Zanfir and Mihai \cite{zanfir2013moving} is used to extract the frames for recognition. The sliding window is simple and easy for use, but its limitation is the fixed scale of the sliding window, since the temporal length of the action is unknown and different due to the variety of the action. In addition, the sliding window method can lose long-term context information, such that it has very low accuracy for such situations. Zolfaghari et al.\cite{zolfaghari2018}  proposed a sampling mechanism for online high-quality action recognition using RGB video stream, which exploits that neighboring frames are largely redundant. In this paper, we use a similar idea for online human action recognition by sampling the kinetic skeleton sequences to balance the data which is far away from the present and the data which is relatively close. We also show that the proposed sampling method can be better than the sliding window for long term actions recognition.
 
\begin{table*}[!t]
	\newcommand{\tabincell}[2]{\begin{tabular}{@{}#1@{}}#2\end{tabular}}  
	\scriptsize
	\centering
	\caption{Geometric feature calculation methods and feature description}
	\label{Tab1}
	\begin{tabular}{m{2.5cm}m{1.6cm}m{4.cm}m{5.cm}}
		\toprule
		\textbf{Feature}&\textbf{Symbol}&\textbf{Calculation Methods}&\textbf{Description} \\
		\midrule
	    Joint Orientation & 
	    $gg\_o$ & 
	    \tabincell{c}{$ gg\_o(g_1,g_2)$= \\ $unit(\stackrel{\longrightarrow}{g_1g_2}) $} & 
	    Direction from joint $ g_1 $ to $ g_2 $  \\
	    
	    \midrule
		Joint Line Distance &
		$ gx\_d $ &  
		\tabincell{c}{$ gx\_d(g,x_{g_1g_2})= $ \\ $2S_{\triangle{gg_1g_2}}/{\left\|\stackrel{\longrightarrow}{g_1g_2} \right\| }$} & 
		Distance from joint g to line $ x_{g_1g_2}$\\
		
		\midrule
		Line Line Angle & 
		$xx\_a$ &
		\tabincell{c}{$xx\_a(x_{g_1g_2},x_{g_3 g_4})=$ \\ 
		$\arccos(gg\_o(g_1,g_2)^T $ \\
		$\bigodot gg\_o(g_3,g_4) $ } & 
		Angle between line $x_{g_1g_2}$ and $x_{g_3 g_4} $\\
		
		\midrule
		Joint Plane Distance & 
		$gP\_d$ &
		\tabincell{c}{$gP\_d(g,P_{g_1g_2g_3})=$ \\ 
		$ (g-g_1) \bigodot gg\_o(g_1,g_2) $\\
		$ \bigotimes gg\_o(g_3,g_4) $ } & 
		Distance from joint g to plane $ P_{g_1g_2g_3} $ \\
		
		\midrule
		Line Plane Angle & 
		$xP\_a$ &
		\tabincell{c}{$xP\_a(x_{g_1g_2},P_{g_3g_4g_5})$ \\ 
		$ =\arccos(gg\_o(g_1,g_2)) \bigodot  $\\
		$ gg\_o(g_3,g_4) \bigotimes gg\_o(g_3,g_5) $ } & 
		Angle between line $ x_{g_1g_2} $ and normal vector of plane $P_{g_3g_4g_5}$   \\
		
	    \midrule
		Plane Plane Angle & 
		$PP\_a$ &
		\tabincell{c}{$PP\_a(P_{g_1g_2g_3},P_{g_4g_5g_6})$ \\ 
		$ =\arccos(gg\_o(g_1,g_2)) \bigotimes  $\\
		$ gg\_o(g_1,g_3) \bigodot gg\_o(g_3,g_4) $ \\
		$ \bigotimes gg\_o(g_3,g_5) $ } & 
		Angle between normal vector of plane $ P_{g_1g_2g_3} $ and normal vector of plane $P_{g_4g_5g_6}$  \\
		\bottomrule
	\end{tabular}
\end{table*}

\section{METHODS}\label{sec:method}

In this section, we introduce our ideas to extract advanced features from original skeleton sequences for handling the problems causing by the view changes, and discuss the idea of memory group sampling to extract the frames from previous frames for handling the problem of unknown starting point of the actions. Finally, a convolution neural network (CNN) is used for action recognition. The overall flowchart of the propose method can be seen in Fig.\ref{fig:flowchart}.

\subsection{Advanced Kinetic Skeleton Feature Representation}\label{AA}
To fully describe the action of human, we use not only the advanced spatial geometrical information of the human joints, but also the temporal motion features.

Joint collection distances (JCD) feature is first proposed in \cite{yang2019make}, which is a location viewpoint invariant feature. If each human skeleton has N joints with corresponding Cartesian coordinates $g^k_i = (x, y, z)$ for the $k_{th}$ frame and the $i_{th}$ joint, the joint collection distances can be calculated as follows:  

\begin{equation}
F^k = \left[
\begin{matrix}
\left\| g^k_2g^k_1 \right\|  \\
\vdots & \ddots\\
\vdots & \cdots & \ddots\\
\left\| g^k_Ng^k_1 \right\| & \cdots & \cdots & \left\| g^k_{N}g^k_{N-1} \right\| 
\end{matrix}
\right]\\
\label{equ:JCD}
\end{equation}
where $ \left\| g^k_ig^k_j \right\| $ represents the Euclidean distance between joint $g^k_i$ and joint $g^k_j$. Since $F^k$ is a symmetry matrix, we only use the lower triangular matrix as JCD features.

In addition to the JCD features, we also explore the feature information from joint orientation, joint-line distance, line-line angle, joint-plane distance, line-plane angle, plane-plane angle from original skeleton joints according the work presented in \cite{zhang2017geometric}. To reduce the information redundancy, we select these lines and planes according to the following rules:

\begin{itemize}
	\item Lines: $ x_{g_1g_2} $ is a line connected by joint $ g_1 $ and $ g_2 $, which satisfies one of the following constraints.
	1.  $ g_1 $ and $ g_2 $ are directly adjacent in the human structure.
	2. One of $ g_1 $ and $ g_2 $ is the end joint (like head joint, left or right hand joint, left or right foot joint), and the other is the joint separated by a joint in the human structure.
	3. $ g_1 $ and $ g_2 $ are both end joints.
	\item Planes: $ P_{g_1g_2g_3} $ is a plane determined by a triangle formed by $ g_1 $, $ g_2 $ and $ g_3 $. Only five planes that corresponding to body, two arms and two legs are considered.
\end{itemize}

\begin{figure}[!t]
	\centering
	\begin{minipage}{2.6cm}
		\includegraphics[width=2.5cm]{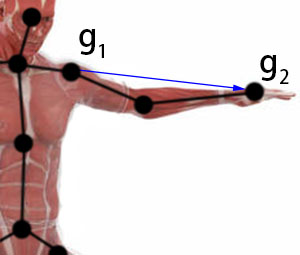}\\
		\centerline{$gg\_o(g_1,g_2)$}
	\end{minipage}
	\begin{minipage}{2.6cm}
		\includegraphics[width=2.5cm]{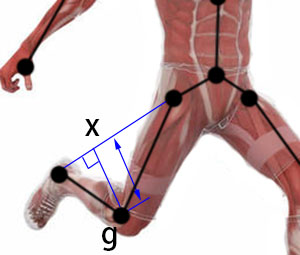}\\
		\centerline{$gx\_d(g,x)$}
	\end{minipage}
	\begin{minipage}{2.6cm}
		\includegraphics[width=2.5cm]{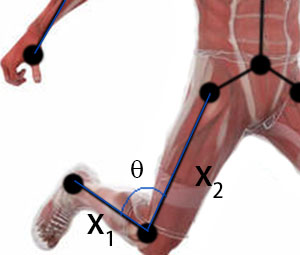}\\
		\centerline{$xx\_a(x_1,x_2)$}
	\end{minipage}
	\begin{minipage}{2.6cm}
		\includegraphics[width=2.5cm]{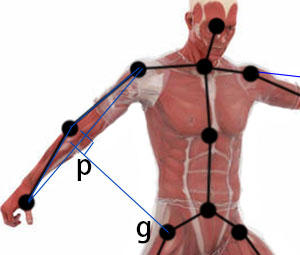}\\
		\centerline{$gP\_d(g,P)$}
	\end{minipage}
	\begin{minipage}{2.6cm}
		\includegraphics[width=2.5cm]{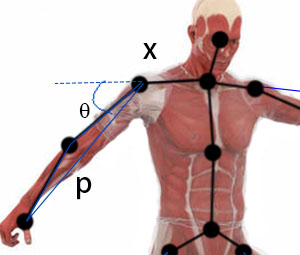}\\
		\centerline{$xP\_a(x,P)$}
	\end{minipage}
	\begin{minipage}{2.6cm}
		\includegraphics[width=2.5cm]{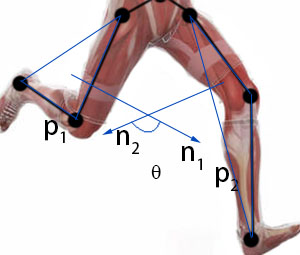}\\
		\centerline{$PP\_a(P_1,P_2)$ }
	\end{minipage}	
	\caption{Six advanced geometric features used in this paper.}
	\label{fig:feature}
\end{figure}

According to these selected lines and planes, six types of geometric features are chosen as shown in the table \ref{Tab1}. Here, we remove these repetitive  features caused by symmetry of the human body. The examples of geometric features are shown in Fig.\ref{fig:feature}.

The geometrical features only depict the spatial relations of the human skeleton joints, whereas the temporal information is missed which is important for human action recognition using a video sequence. Therefore, we further employ the global motion features by differential the spatial position $ G^{k}$ of human skeleton joints between two $k_{th}$ frame and $k_{th}+s$ frame where $s$ is the temporal scale. Here we use two scales for capture the fast motion  $Y^k_{fast} $ and slow motion $Y^k_{slow}$, i.e., $s=1, 2$. The motion features are calculated as

\begin{equation}
Y^k_{slow} = G^{k+1}-G^k , k \in \left\{1,2,\cdots,K-1 \right\} 
\label{equ:sm}
\end{equation}

\begin{equation}
Y^k_{fast} = G^{k+2}-G^k , k \in \left\{1,2,\cdots,K-2 \right\} 
\label{equ:fm}
\end{equation}

\subsection{Memory Group Sampling Mechanism}

For online action recognition, the unknown start and end action time is a challenge problem compared to the offline action recognition which uses segmented action sequences. The sliding window method is the popular method for tradition online action recognition \cite{zanfir2013moving}, which has fixed window size and can lose long-term context information. We here propose a memory group sampling mechanism to balance the information that near and far from current frame. A fixed number of frames are chosen as the input of the action classifier. The sampling function is defined as:

\begin{equation}
I^T = \left\{ 0.5^TQ^0 \right\} \bigcup_{t=1}^T \left\{ 0.5^{T-t+1}Q^t \right\}    
\label{sample1}
\end{equation}

\begin{equation}
T = \lfloor \frac{j}{N} \rfloor - 1  
\label{sample2}
\end{equation}
where $I^T$ is the image frames obtained in the T-th sampling, $ Q^t $ is a queue which stores $N$ consecutive frames of data before the sampling step t in the data stream, $j$ is the number of frames currently received, $0.5$ means $50\%$ sampling of the data. $N$ is the number of sampling frames required for action classifier input. 

At the beginning of the video sequence (T=0), we use all N frames of data received in the current data stream:
\begin{equation}
I^0 = \left\{ 0.5^TQ^0 \right\} = Q^0
\label{sample3}
\end{equation}
For the third sampling (T=2), the sampling equation is shown as:
\begin{equation}
I^2 = \left\{ 0.5^2Q^0 \right\} \bigcup \left\{ 0.5^2Q^1\right\} \bigcup \left\{ 0.5^1Q^2\right\}   
\label{sample4}
\end{equation}
where $ I^2 $ consists of three parts, including $25\%$ of $Q^0$, $25\%$ of $Q^1$ and $50\%$ of $Q^2$. It shows that the latest frames has more probability to be chosen than the older frames, whereas the long-term contextual information is also considered. The specific steps of the memory group sampling algorithm are shown in Algorithm \ref{alg:mem}.

To store these sampled frames, we use a memory group $M$ which will be replaced by the working group $I$ after each sampling step, such that the $I$ can be defined as 
\begin{equation}
I^T=\begin{cases}	Q^0    &\mbox{if T = 0 }\\
	\left\{ 0.5Q^t \right\} \bigcup \left\{ 0.5M\right\}  &\mbox{if T $>$ 0}
\end{cases}
\label{sample5}
\end{equation}
The update of $M$ ensures that the frame at a closer time has a greater sampling density. Thereby, it has greater weight when inputting the model, which can solve the problem of sliding window. After the sampled data is input into classifier at each time step to make real-time prediction, the prediction result at the current moment and previous are added and averaged to obtain the final prediction result.

\begin{algorithm}[!t]
	\caption{Online Action Recognition Based on Memory Group Sampling Mechanism}
	\label{alg:mem}
	\begin{algorithmic}[1]
		\Require Live human skeleton stream $L$, trained classifier $C$, sampling frame number $N$.
		\State {Initialize the empty queue $Q$ to save the sampled N frames}
		\State {Initialize the memory group $M$}
		\State {Initialize output average recognition probability $p_a$}
		\label{code:recentStart}  
		\While {New frame available from $L$}  
		\State {Add frame $f_i$ to queue $Q$}
	    \If {$i\%N$} 
	    \State {$I$ = sample $50\%$ $Q$ and sample $50\%$  $M$}
	    \State {Empty queue $Q$}
	    \State {Feed $I$ to the classifier $C$ to get recognition probability  $p$}
	    \State {$M = I$}
	    \State {$p_a$ = average $p_a$ and $p$}
	    \State {Output action recognition probability $p_a$}
	    \EndIf

		\EndWhile  
		\label{code:recentEnd}  
	\end{algorithmic}
\end{algorithm}

\subsection{1D CNN for Online Action Recognition }

\begin{figure}
\centering
\includegraphics[width= 0.48\textwidth]{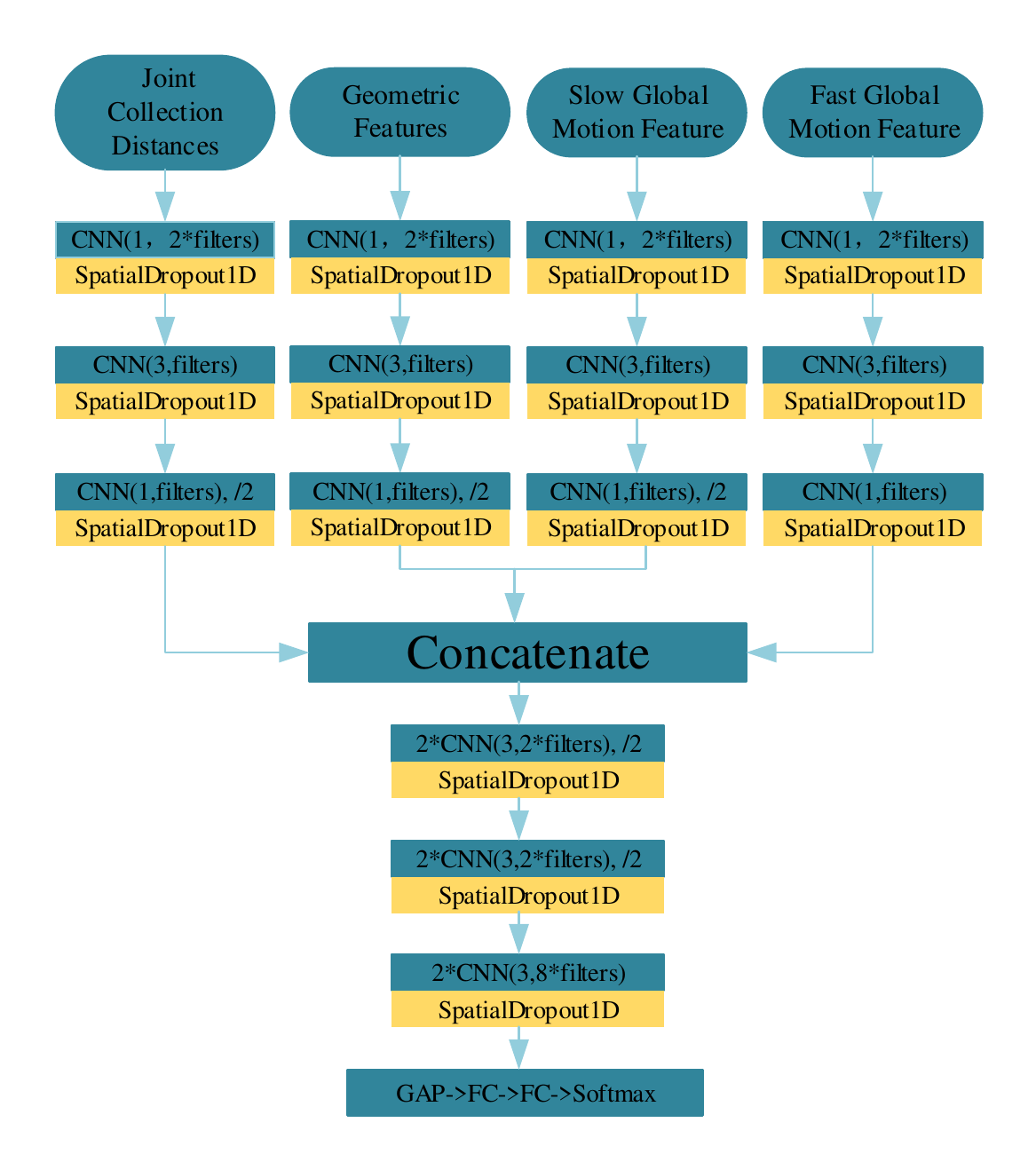}\\
\caption{The network architecture of our methods. “2×CNN(3, 2*filters), /2” indicates two 1D CNN layers (kernel size = 3, channels = 2*filters) and a Maxpooling (strides = 2). “SpatialDropout1D” indicates one 1D space dropout layer for suppressing overfitting. GAP indicates Global Average Pooling. FC indicates Fully Connected
	Layers. }
\label{fig:net}
\end{figure}

After obtaining geometrical feature and motion feature representations of the original data, we concatenate all features to a vector, and use a 1D convolution neural network (CNN) \cite{yang2019make} to train the action classifier by using all features. The architecture of the 1D CNN is shown in Fig. \ref{fig:net}. This 1D CNN classier is very fast and efficient, which is suitable for online action recognition. 

\section{EXPERIMENTS}\label{sec:exp}

\begin{table}[!t]
	\scriptsize
	\centering
	\caption{Offline Action Recognition Experiments on JHMDB Dataset}
	\label{Tab2}
	\begin{tabular}{m{6 cm}m{1.5cm}}
		\toprule
		Methods & Accuracy(\%) \\
		\midrule
		ChainedNet (ICCV17) \cite{8237578} & 56.8\\
		EHPI (ITSC19) \cite{8917128} & 65.5 \\
		PoTion (CVPR18) \cite{8578832} & 67.9\\
		DDNet(2019) \cite{yang2019make} & 77.2 \\
		\textbf{Ours} & \textbf{78.5} \\
		\bottomrule
	\end{tabular}
\end{table}

\begin{table}[!t]
	\scriptsize
	\centering
	\caption{Offline Action Recognition Experiments on UT-Kinect Dataset}
	\label{Tab3}
	\begin{tabular}{m{6 cm}m{1.5cm}}
		\toprule
		Methods & Accuracy(\%) \\
		\midrule
		SkeletonJointFeatures(2013) \cite{6595918} & 87.9 \\
		ElasticFunctionalCoding (2015) \cite{7298934} & 94.9 \\
		GeoFeat(2017) \cite{zhang2017geometric}& 95.9\\
		GFT (2019) \cite{8803186} & 96.0 \\
		\textbf{Ours}  & \textbf{96.9} \\
		\bottomrule
	\end{tabular}
\end{table}

\begin{table*}[!t]
	\scriptsize
	\centering
	\caption{Online Action Recognition Experiments on UT-Kinect Dataset}
	\label{Tab4}
	\begin{tabular}{lllllllllllll}
		\toprule
		& Walk & Sit down & Stand up & Pick up & Carry & Throw & Push & Pull & Wave & Clap hands &Average\\
		\midrule
		Memory group sampling  & 100\% & 100\% & 100\% & 100\% & 78\% & 70\% & 80\% & 90\% & 100\% & 100\% &92\% \\
		Sliding window & 30\% & 30\% & 80\% & 100\% & 78\% & 70\% & 80\% & 90\% & 100\% & 90\% &74.7\% \\
		\bottomrule
	\end{tabular}
\end{table*}

To demonstrate the performance of the proposed method, 
we use two public skeleton-based datasets: JHMDB dataset \cite{JHMDB} and UT-Kinect dataset \cite{UTKinect}. The JHMDB dataset contains 928 sample data with 2D skeleton data inferred from RGB data. The UT-Kinect dataset contains 200 sequences of 10 action classes with the 3D skeleton data from depth camera. Every action is recorded twice for each subject. 

To train the neural network, we use a computer with a Nvidia TITAN X GPU. The Adam \cite{kingma2014adam} optimizer is used for learning. We set the initial learning rate to 0.001. When the loss function value of the validation set does not decrease after more than 5 epochs of training, the learning rate is reduced at a rate of 0.5 until it is reduced to 0.00001. A total of 400 epochs were trained.

To show the advance performance of the proposed spatial and temporal kinetic skeleton features, we compare our method to the state of the art action recognition methods in the offline manner using JHMDB and UT-Kinect datasets respectively, which is shown in Table. \ref{Tab2} and Table. \ref{Tab3}.

Our algorithm achieves higher classification accuracy than ChainedNet, EHPI, PoTion and DDNet on JHMDB dataset. Compared to our method, ChainedNet \cite{8237578} uses a 3D CNN classifier and directly inputs the original joint sequences, which achieves $56.8\%$ recognition rate. Similarly, EHPI \cite{8917128} encodes original joint coordinates as the color information over a fixed period of time, such that the motion of joints can be seen from color changes, then this color image is fed into a CNN for classification, which achieves $65.5\%$ recognition rate.  PoTion \cite{8578832} extracts joint heatmaps for each frame and colorize them using a color that depends on the relative time in the video clip. For each joint, they aggregate them across all frames, which constitutes coded images that are further stacked together as an action representation for classification, which achieves $67.9\%$ recognition rate. DDNet \cite{yang2019make} employs 1D CNN network for action classification using the JCD feature and global motion feature, which achieve less accuracy result $77.2\%$ since our method explores more advanced geometrical features from original skeleton data as shown in Fig. \ref{fig:feature}. 

For UT-Kinect dataset, we use half of the samples for training and the other half for testing. Our methods achieve the competitive performance compared to the state of the art methods. Skeleton Joint Features based method proposed by \cite{6595918} computes the frame difference and pairwise distance of skeleton joints positions to characterize the spatial information of the joints in 3D space, which achieves $87.9\%$ recognition accuracy. Elastic Functional Coding based method proposed by \cite{7298934} employs the TSRVF space that provides an elastic metric between two trajectories on a manifold to learn the latent variable space of human actions, and propose mfPCA for compact and robust representation of features, which achieves $94.9\%$ recognition accuracy. GeoFeat \cite{zhang2017geometric} uses advanced geometric features to model human action sequences, and uses a three-layer LSTM network to classify actions, which achieves $95.9\%$ recognition accuracy. GFT \cite{8803186} leverages skeletal temporal graph structures to represent body joints, and the graph transform GFT is utilized to extract representations of human motion data, which achieves $96\%$ recognition accuracy. Compared to these methods, we use advanced geometric features and multi-scale motion features to capture spatial and temporal information of human action, and employ a 1D CNN for action classification, which achieves $96.9\%$ recognition accuracy.

\begin{figure}[!t]
\centering
\includegraphics[width=0.48\textwidth]{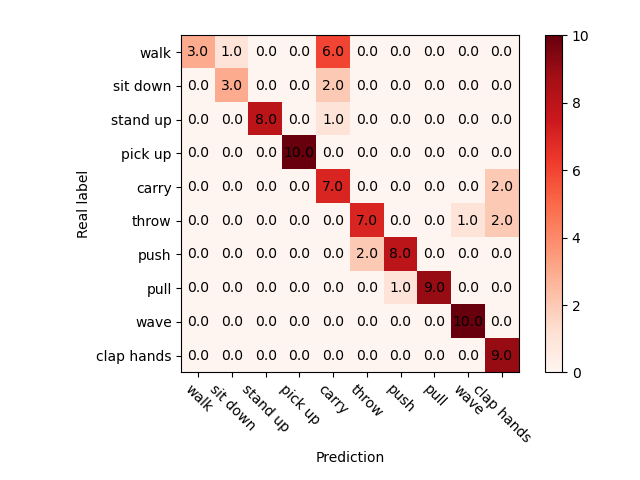}\\
\caption{The confusion matrix of sliding window based recognition on UT-Kinect dataset. }
\label{fig:conf-sw}
\end{figure}

\begin{figure}[!t]
\centering
\includegraphics[width=0.48\textwidth]{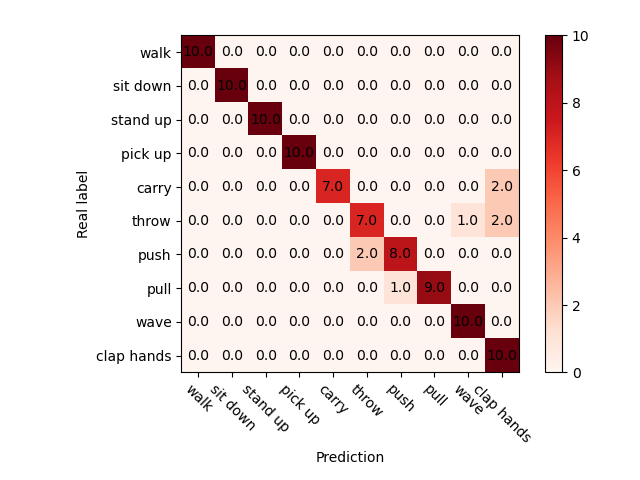}\\
\caption{The confusion matrix of memory group sampling based recognition on UT-Kinect dataset. }
\label{fig:conf-mgs}
\end{figure}

To show the effectiveness of the memory group sampling for online action recognition using skeleton sequences, we compare the proposed method to the sliding window method. We first define online action recognition rate as the ratio of the number of positive samples to the total number of samples. The size of the input samples for the online action classifiers is $16$, so the sliding window has a fixed size as shown in \cite{zanfir2013moving}. The results are shown in Table. \ref{Tab4}, which shows that memory group sampling achieves higher mean accuracy. For short action sequence, the sliding window method has similar performance with ours, such as the action pick up, carry, throw, push, pull, wave. However, the sliding window can not handle the long action sequences well, such as walk, sit down, stand up and clap hands due to its fixed window size. The confusion matrices of human action recognition using sliding window and memory group sampling are shown in Fig. \ref{fig:conf-sw} and Fig. \ref{fig:conf-mgs} respectively.

\section{CONCLUSIONS} \label{sec:conc}

In order to solve the problem of online human action recognition, we propose a memory group sampling based 1D CNN action classifier using both of the spatial and temporal kinetic skeleton features. The memory group sampling is superior to the traditional sliding window method, since it can capture long term context information while the nearby frames have higher sampling density. In addition, we combine the JCD feature, advanced geometrical feature and global motion feature to represent the human action information, such that the spatial and temporal information are considered. Furthermore, a simple and effective 1D CNN is used for online classification of concatenated multiple skeleton features. Finally, we demonstrate our method on JHDMB and UT-Kinect datasets, and compare to the state of the art methods, which shows that the proposed method can achieve competitive performance. 


\bibliographystyle{ieeetr}
\IEEEtriggeratref{9}
\bibliography{ref}

\end{document}